\documentclass[10pt,twocolumn,letterpaper]{article}

\usepackage{cvpr}              % To produce the CAMERA-READY version
\usepackage[accsupp]{axessibility}
\usepackage{graphicx}
\usepackage{amsmath}
\usepackage{amssymb}
\usepackage{booktabs}
\usepackage{multicol}
\usepackage{multirow}
\usepackage{amsmath, amssymb, amscd, amsthm, amsfonts}
\usepackage[export]{adjustbox}
\usepackage[normalem]{ulem}
\useunder{\uline}{\ul}{}
\usepackage[table]{xcolor}

\usepackage[pagebackref,breaklinks,colorlinks]{hyperref}

\usepackage[capitalize]{cleveref}
\crefname{section}{Sec.}{Secs.}
\Crefname{section}{Section}{Sections}
\Crefname{table}{Table}{Tables}
\crefname{table}{Tab.}{Tabs.}

\def\cvprPaperID{40} % *** Enter the CVPR Paper ID here
\def\confName{CVPR}
\def\confYear{2022}

\begin{document}

\title{DRT: A Lightweight Single Image Deraining Recursive Transformer}

\author{Yuanchu Liang$^1$, Saeed Anwar$^{1,2}$, and Yang Liu$^1$\\
$^1$Australian National University, Australia \quad $^2$Data61-CSIRO, Australia\\
{\tt\small \{yuanchu.liang, saeed.anwar, yang.liu3\}@anu.edu.au}
}

\maketitle

\begin{abstract}
Over parameterization is a common technique in deep learning to help models learn and generalize sufficiently to the given task; nonetheless, this often leads to enormous network structures and consumes considerable computing resources during training. 
Recent powerful transformer-based deep learning models on vision tasks usually have heavy parameters and bear training difficulty.
However, many dense-prediction low-level computer vision tasks, such as rain streak removing, often need to be executed on devices with limited computing power and memory in practice. 
Hence, we introduce a recursive local window-based self-attention structure with residual connections and propose deraining a recursive transformer (DRT), which enjoys the superiority of the transformer but requires a small amount of computing resources.
In particular, through recursive architecture, our proposed model uses only \textbf{$\sim$ 1.3\%} of the number of parameters of the current best performing model in deraining while \textbf{exceeding} the state-of-the-art methods on the Rain100L benchmark by at least \textbf{0.33 dB}. 
Ablation studies also investigate the impact of recursions on derain outcomes. Moreover, since the model contains no deliberate design for deraining, it can also be applied to other image restoration tasks. Our experiment shows that it can achieve competitive results on desnowing.
The source code and pretrained model can be found at \url{https://github.com/YC-Liang/DRT}. 
\end{abstract}  
\section{Introduction}
\label{sec:Intro}

\begin{figure}
\begin{center}
\begin{tabular}{@{}c@{}c@{}c}
  \includegraphics[width=0.15\textwidth]{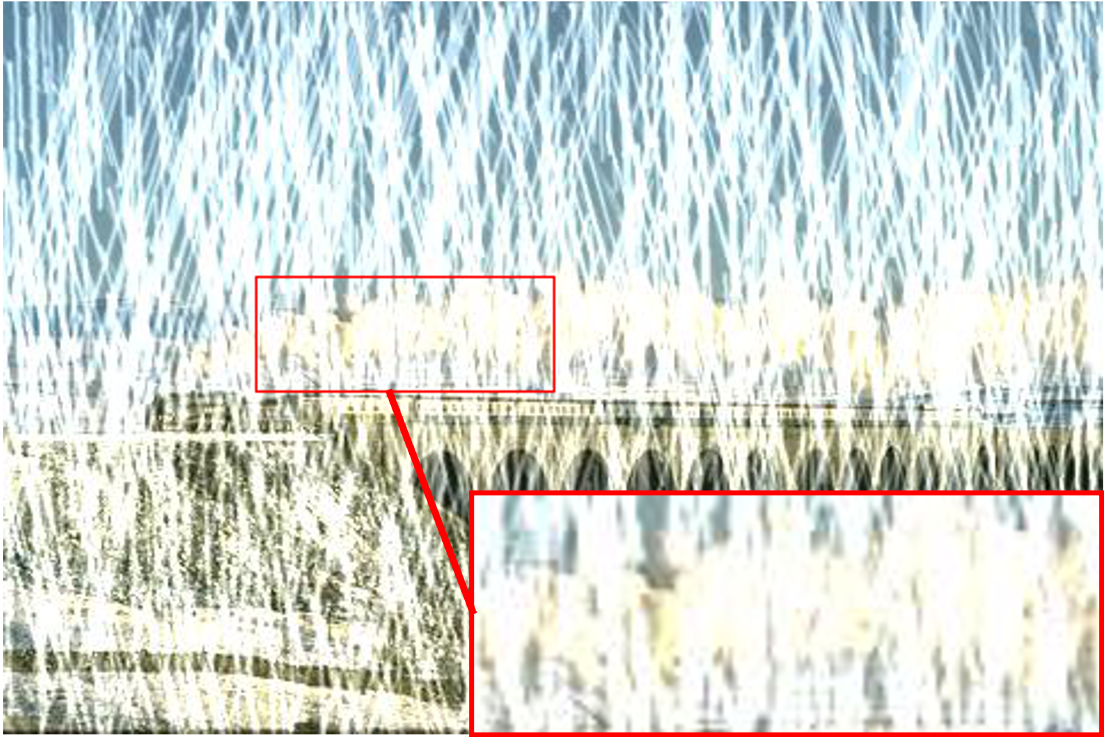} & \includegraphics[width=0.15\textwidth]{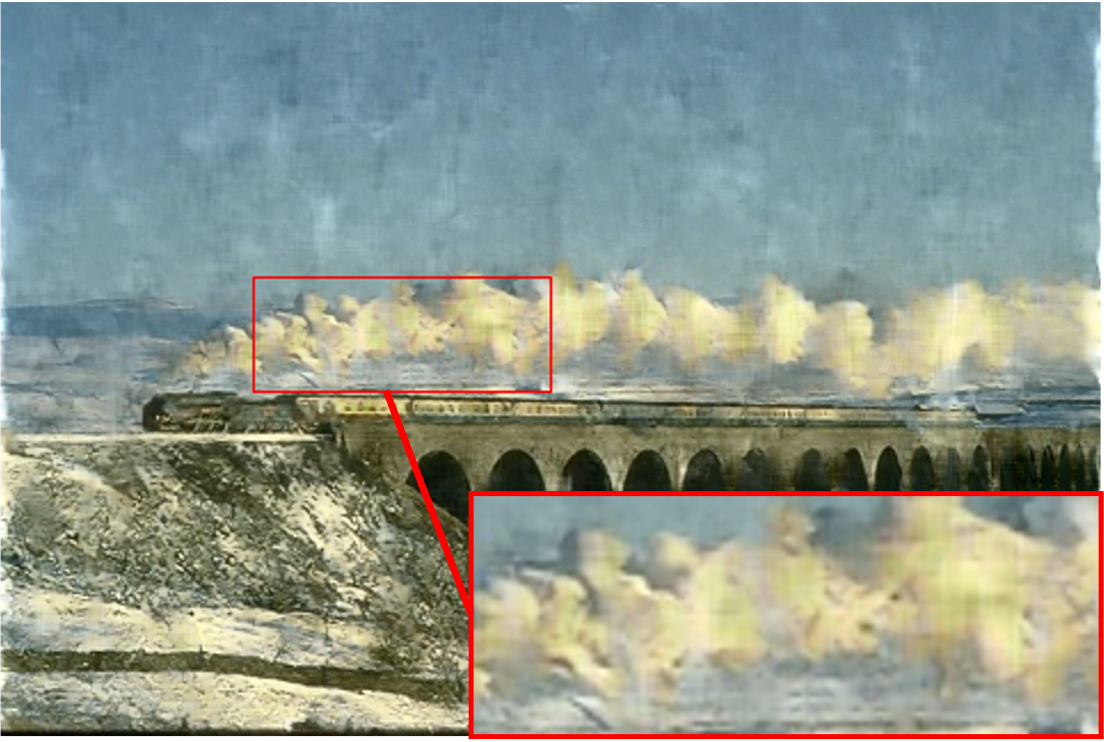}  & \includegraphics[width=0.15\textwidth]{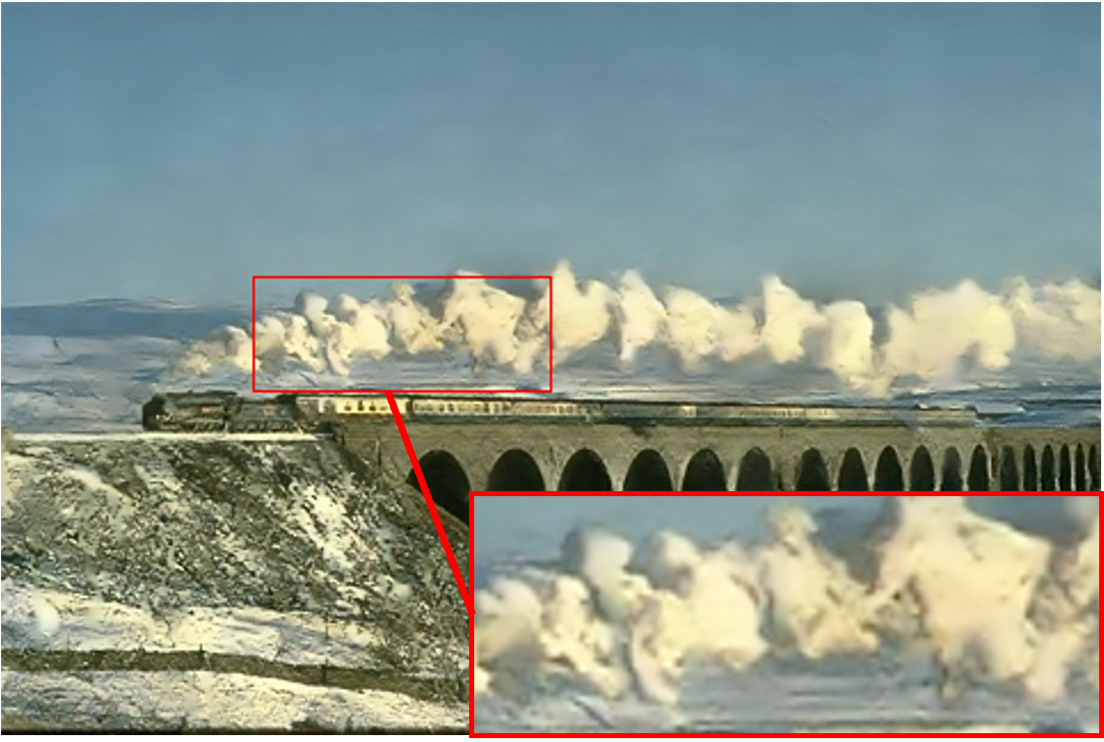} \\
  
  Input & JORDER & PreNet\\
  
  \includegraphics[width=0.15\textwidth]{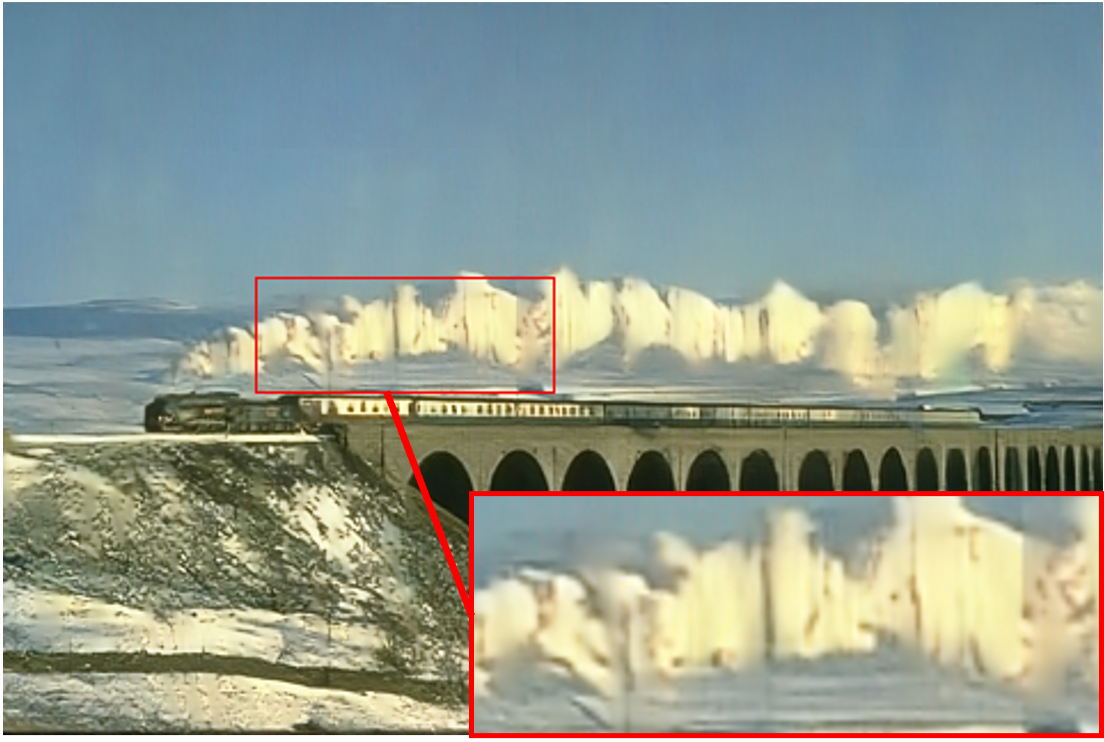} & \includegraphics[width=0.15\textwidth]{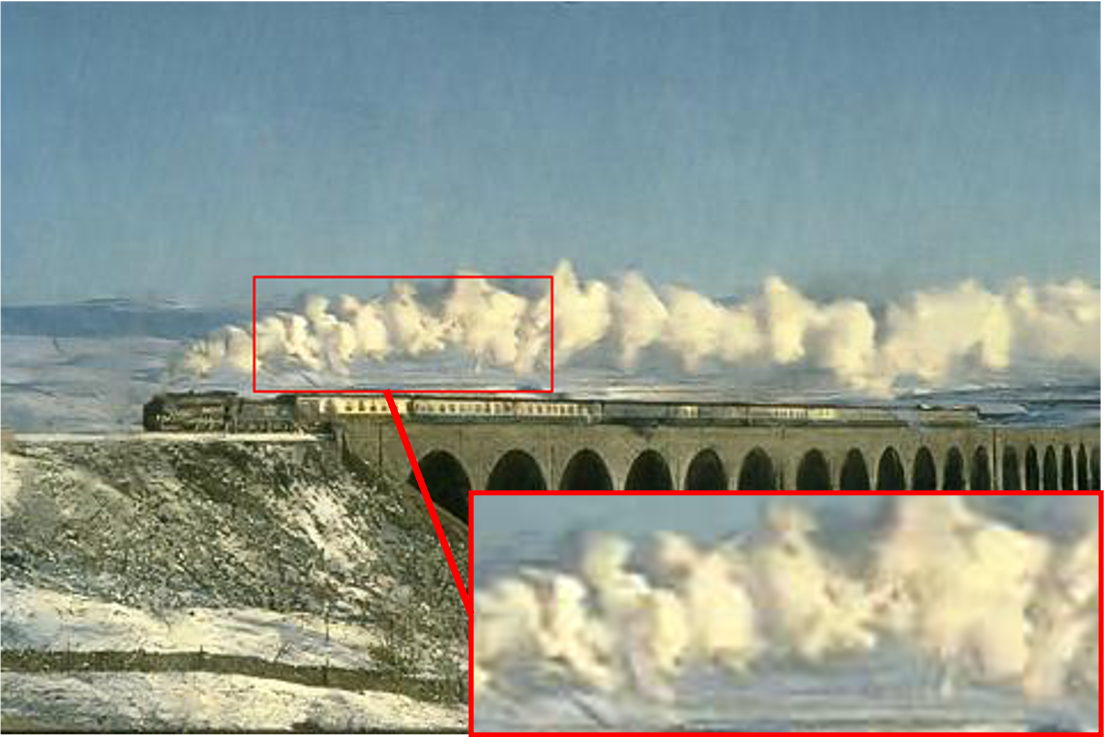}  & \includegraphics[width=0.15\textwidth]{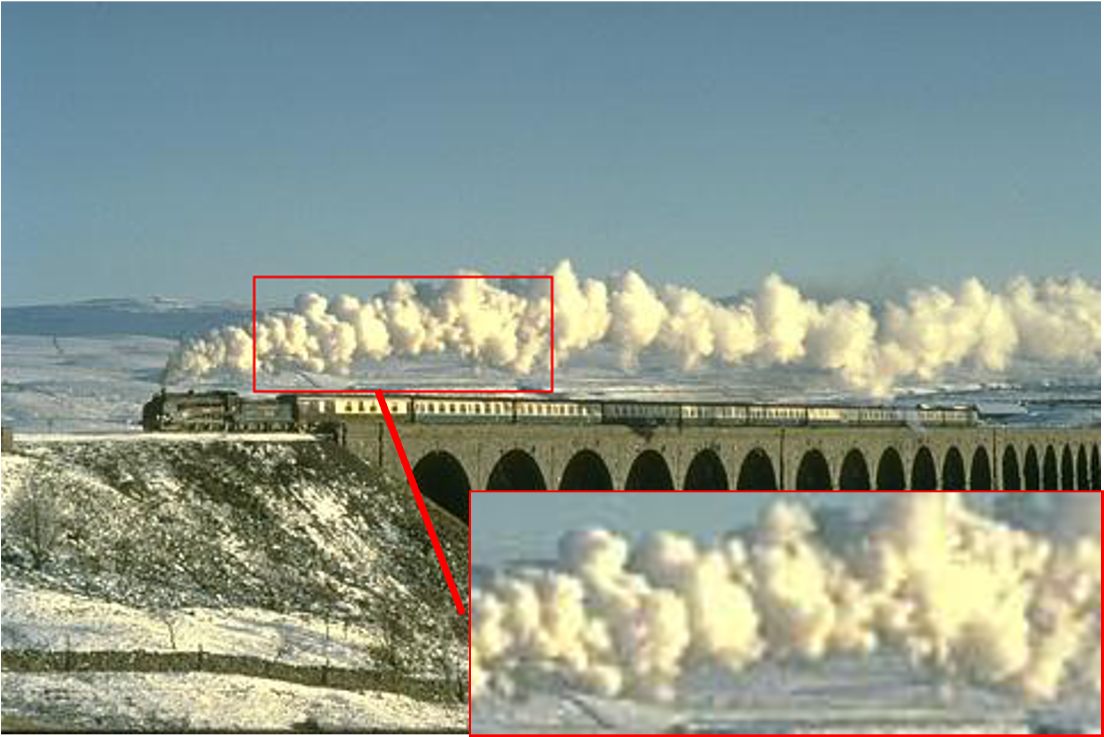} \\

  Hi-Net & DRT (Ours) & GT \\
\end{tabular}
\caption{An example from the Rain100H dataset. Boxed regions are zoomed-in and displayed at the bottom-right corner of each image. Complex irregular shapes, such as smokes produced by the train, are better reconstructed by DRT whereas other methods suffer from artifacts.}
\label{fig:Rain100H}
\vspace{-4mm}
\end{center}
\end{figure}

Computer vision is a fast-growing field and has many real-world applications such as facial recognition for security, object detection for autonomous vehicles, and scene understanding for caption generation. These high-level computer vision tasks often need to operate on non-corrupted images for the best performance. Images taken under rainy weather are naturally corrupted as rain streaks block the background and lead to the loss of information. Hence, finding an efficient way to remove rain streaks from the image and reconstruct the background while keeping the model simple is essential for high-level vision tasks to operate in rainy situations, especially for devices with limited computing resources.

In the last five years, many models have widely adopted Convolutional Neural Networks (CNNs) to perform single image deraining. Specifically, Fu~et~al.~\cite{fu2017clearing} designed an essential lightweight CNN that takes high-frequency components of the image as input to remove rain streaks; the resulting image is added back to the low-frequency part of the image, much like a residual connection. Following this work, much more advanced CNN architecture is employed to achieve better outcomes. Fan~et~al.~\cite{fan2018residual} uses multiple residual connections and recursive CNN blocks to perform rain streak removal. Li~et~al.~\cite{li2018non} utilize non-locally enhanced dense blocks and show that a deep network with a carefully designed structure can enhance the training robustness and achieve better results. Single image deraining methods prior to 2020 have been well summarised by Yang~et~al.~\cite{yang2020single}; we refer to this literature survey for detailed information. Recently, many vision tasks have been well performed by architectures based on attention mechanisms proposed by \cite{vaswani2017attention}; however, they can be computationally heavy and time-consuming to train, as discussed in more detail in the next section. This limits the range of domains that vision transformers can apply.

To design an efficient deep vision transformer for image reconstruction tasks, we propose a recursive attention structure with skip-connection, termed Deraining Recursive Transformer (DRT). In particular, a well-designed recursive structure shares weights among certain self-attention blocks from the network, so the number of parameters no longer increases dramatically while increasing the network's depth. Moreover, in image deraining, the input image and output image have a large portion of overlapping information; therefore, by leveraging residual connections from the input to the output, a small number of self-attention blocks are needed to focus on detecting rain streaks and hence reducing the size of the network. With these designs, our model only needs $\sim$ 1.18M parameters (i.e., $\sim$ 1.3 \%  of the current state of the art in single image deraining) to outperform other derain methods on the Rain100L benchmark by at least 0.33dB. An example of derain outcomes of different methods is displayed in \ref{fig:Rain100H}, which shows the proposed DRT has a strong ability to reconstruct various irregular backgrounds. Furthermore, unlike ViT proposed by \cite{dosovitskiy2020image} or IPT proposed by \cite{chen2021pre}, which are trained on millions of images, DRT only needs to be easily trained on a small data set (700 images) and hence effectively reduces the training difficulty. In summary, this work makes the following contributions, 

\begin{itemize}
    \item To the best of our knowledge, this is the first work investigating recursively structured vision transformers for image reconstruction tasks.
    \item Our model outperforms many others methods on the task of single image deraining while consuming less computing resources.
    \item The proposed model is general enough to carry out other image restoration tasks such as single image desnowing.
\end{itemize}

%
%The rest of the paper is organized as follows; Section II will introduce the past techniques relevant to the DRT; %Section III provides the design details of DRT; experimental results and ablation studies are presented in Section IV, %followed by a conclusion in Section V.
\section{Related Work}
\subsection{Vision Transformers}
Since the introduction of the transformer network by Vaswani et al. \cite{vaswani2017attention}, much research has been done to replace CNNs with transformers as the backbone structure for vision tasks. In particular, Dosovitskiy et al. \cite{dosovitskiy2020image} introduce the Vision Transformer (ViT) by splitting an image into non-overlapping small patches and treating each of the patches as a token in the original transformer. ViT performs well on the classification tasks while maintaining a relatively low computational complexity due to the image patch embedding mechanism. To extend the capability of ViT, Liu et al. \cite{liu2021swin} propose the Swin-Transformer. One of the purposes behind the design of the Swin transformer is the same as the purpose of our work, to reduce the computational complexity of the vision transformer. This type of transformer performs multi-headed self-attention in local windows instead of globally as in ViT. The computation cost is reduced from $4hwC^2 + 2(hw)^2C$ to $4hwC^2+2(M)^2hwC$, where the image dimension is $C\times h\times w$, and the local window size is denoted by $M$.

Meanwhile, Swin-Transformer employs the shifted window approach to compensate for the loss of global information. A single Swin Transformer Block (STB) firstly passes the input through a layer norm (LN) and then via a windowed multi-headed self-attention (WMSA) mechanism. The output from the self-attention mechanism is combined with the initial input forming the first residual connection. Then the output is passed through a multi-layer perceptron (MLP) followed by another LN. The final output combines again with the output from the self-attention mechanism forming the second residual connection. Multiple STBs are used to process a given image and achieve state-of-the-art results in recognition, detection, and segmentation tasks.

%%% placed the network image here for better formatting 
\begin{figure*}[t]
    \centering
    \includegraphics[width=\textwidth]{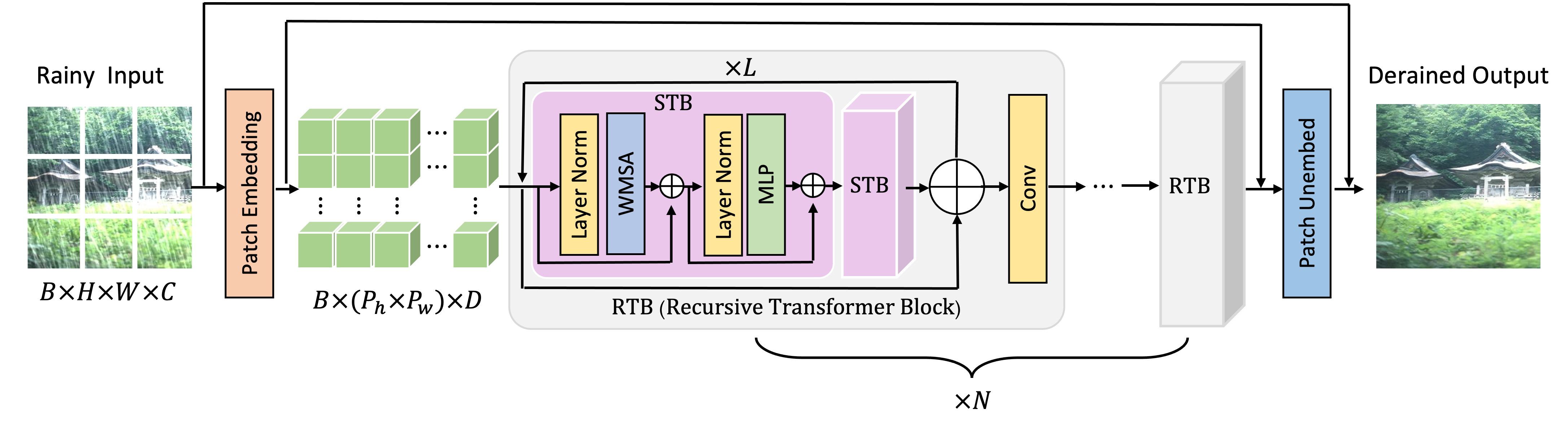}
    % \vspace{-2mm}
    \caption{The Deraining Recursive Transformer architecture. RTB stands for recursive transformer block, and STB stands for Swin Transformer block. N refer to the number of RTBs and L refers to the number of recursive calls. As mentioned in the ablation study section, the number of RTBs, STBs and recursive depth can be made arbitrarily large.}
    \label{fig:DRT}
\end{figure*}
%%%

\subsection{Applications of Vision Transformers to Image Reconstruction}
Vision transformers from the last section and their variants have been utilized recently to achieve better general image restoration outcomes. Chen et al.\cite{chen2021pre} uses a transformer encoder-decoder structure based on \cite{vaswani2017attention} to design a general image restoration transformer termed Image Processing Transformer (IPT). Similarly, Wang et al. \cite{wang2021uformer} propose to use a transformer block, namely LeWin block, and an hourglass structure with skip-connections for background reconstructions. This design performs well in different image reconstruction tasks while maintaining a relatively low computational cost. Furthermore, Liang et al. \cite{liang2021swinir} design residual transformer blocks on top of the Swin-Transformer and stack multiple of the residual blocks to perform deep feature extractions for image reconstructions. Unlike IPT, which is pre-trained on ImageNet, this type of transformer is trained with 800 samples while maintaining a small number of network parameters relative to many other CNN-based or transformer-based methods.

\subsection{Recursive Network Structures}
A recursive structure in deep learning repetitively calls some portion of layers in a network to process data. The use of the recursive structure for image reconstruction tasks can be traced back to \cite{kim2016deeply}. In this work, a deeply recursive CNN has been proposed to stabilize the training process and maintain a deep structure with few parameters. It empirically shows that a single recursive layer is enough to achieve good image super-resolution results. Later, the same idea is used by \cite{fan2018residual} to design a residual guided recursive CNN network for image deraining. Our method is similar to this work because we also thoroughly investigate the use of recursive and residual structures on transformer-based architectures instead of relying on CNNs. Although recursive structures can also lead to potential issues such as exploding gradient or overfitting, carefully designed recursions can lead to optimal network architecture, as discussed in the following sections. 
\section{Method}
\subsection{Network Architecture}
DRT consists of three stages, the patch embedding stage, $f_1$, the deep feature extraction stage $f_2$, and the image reconstruction stage, $f_3$. Its visualisation is given in Figure \ref{fig:DRT}. Essentially, one can view the first stage as input pre-processing stage and the third stage as inference output post-processing stage whereas the second stage is the main inference stage. Given a raw rainy image $R_{\text{input}}$, the equation below hence captures the functionality of DRT,
\begin{equation}
    \text{DRT}(R_\text{input}) = f_3\circ f_2 \circ f_1(R_{\text{input}})
\end{equation}

The patch embedding stage $f_1$ firstly applies a convolutional layer to the rainy image $R_\text{input} \in \mathbb{R}^{H \times W \times C}$ and map it to $R_\text{conv} \in \mathbb{R}^{H//P \times W//P \times D}$ to divide the image into patches and stack them depth-wisely. Here, $H\times W\times C$ is the original image's height, width and depth respectively. $P$ denotes the patch size used to divide the input into small non-overlapping windows, and $D$ is the embedded dimension and usually $D \gg C$. Note the $//$ sign denotes the integer division operation. Since transformers typically work with one dimensional tokens, we reshape the convolution layer's output $R_\text{conv} \mapsto R_\text{embed} \in \mathbb{R}^{HW//P \times D}$ to complete the first stage of the network. This output can then be fed to transformer blocks to process. We can succinctly write the first stage as
\begin{equation}
    f_1(R_{input}) = \text{PatchEmbed}(\text{Conv}(R_{input})).
\end{equation}

The next stage consists of multiples of recursive transformer blocks (RTB) composed together to perform deep feature extractions as described by Equation \ref{f2}. We use $N$ to denote the total number of RTBs. The details of the RTB are given in the next subsection. Here we note that this stage maintains the input dimension, and there is no parameter sharing between each RTB. The output from this stage is denoted as $R_\text{deep}$.  At the end of the process, a residual connection is used to add the input to the deep feature extraction stage to its output to restrict the RTBs from focusing on detecting rain streaks.
\begin{equation}
\label{f2}
    f_2(R_{\text{embed}}) = \text{RTB}_1 \circ \dots \circ \text{RTB}_N(R_{\text{embed}}) + R_{\text{embed}}. 
\end{equation}

The deep features, $R_\text{deep}$, are then processed by the image restoration stage, which reverses the process of the first stage as seen in equation \ref{f3}. The height and width of the input image is firstly restored by reshaping $R_\text{deep} \mapsto R_\text{unembed} \in \mathbb{R}^{H//P\times W//P \times D}$ and convolution layers are used to process these features to give the output $R_\text{out} \in \mathbb{R}^{H\times W \times C}$. Finally, another residual connection is used between the network's input to its output so that rain streaks features extracted by the network can be removed from the input.
\begin{equation}
    \label{f3}
    f_3(R_\text{deep}) = \text{Conv}(\text{PatchUnembed}(R_\text{deep})) + R_\text{input}.
\end{equation}

\subsection{The Recursive Transformer Block}
Each RTB uses the recursive and residual connections to stack Swin Transformer Blocks (STBs). Mathematically, the STB can be described as,
\begin{multline}
    STB(x) = MLP \circ LN (WMSA\circ LN(x) + x) \\
    + (WMSA\circ LN(x) + x)
\end{multline}
Here the WMSA can be summarised by the equations \cite{vaswani2017attention},
\begin{equation}
    \text{MultiHead}(Q, K, V) = \text{Concat}(\text{h}_1 \dots \text{h}_n)W^O
\end{equation}
\begin{equation}
    \text{h}_i = \text{Attention}(QW_i^Q, KW_i^K, VW_i^V)
\end{equation}
\begin{equation}
    \text{Attention}(Q, K, V) = \text{Softmax}\left(\frac{QK^T}{\sqrt{d_k}}\right)V
\end{equation}
In the above equations, $W$ refers to weights. The inputs, Q, K, and V, often are called queries, keys, and values. The term $\sqrt{d_k}$ refers to the dimension of $K$. If we denote the number of STB in each residual connection as $U$ and the number of recursive calls as $L$, let the input be denoted as $S_\text{in} \in \mathbb{R}^{HW//P \times D}$, then each RTB can be described with the following equation.
\begin{equation}
RTB(S_\text{in}) = \bigoplus_{j=1}^L \left(\bigoplus_{i=1}^U STB_i(S_\text{in}) + S_\text{in}\right).
\end{equation}

The $\bigoplus$ sign is used to denote multiple function compositions. The dimension of the input and output is kept the same. The residual connection always starts from the input to the RTB. The first $\bigoplus$ in the above equation controls the recursive calls of RTB, and the second $\bigoplus$ controls the composition of different STBs. Each $STB_i$ in a residual connection can have different parameters, but recursive calls always invoke the same STBs as seen in the equation above since the index $j$ is never used. For instance, in Figure 2, there are two STBs ($U=2$) in each residual connection, and such connection is recursively called three times ($L=3$) in a single RTB. At the end of each RTB, convolutional layers are used to process information that the local window-based Swin transformer may ignore. If a single convolutional layer is employed, we do not use any activation after the convolution. For multiple convolutional layers, we use Leaky ReLU as activation functions.

\subsection{Loss Function}
There are many proposed loss functions to facilitate the training of the networks; however, we find it is sufficient to use the mean squared error loss. For an output image, $x$ and its corresponding ground-truth image $\bar{x}$, the error is given by,
\begin{equation}
    \text{MSE}(x) = \frac{1}{M} \sum_{i=1}^M (\bar{x}_i - x_i)^2,
\end{equation}
where M denotes the total number of entries, $\bar{x}_i$ denotes the ground truth pixel value of the i$^{th}$ entry, and $x_i$ denotes the prediction value of the i$^{th}$ entry. For multiple predictions, we simply average the errors for each pair to get a final error measure.
\section{Experiments}
\begin{table*}[t]
\caption{Derain results of different methods on three data sets evaluated using the PSNR and SSIM metric. The number of parameters used by each method is displayed on the right-most column. Older methods that did not publish code are ignored from this column. Our results are displayed in bold.}
\begin{center}
%\rowcolors{3}{gray!25}{white}
%\resizebox{15cm}{!}{
\begin{tabular}{|l|cc|cc|cc|c|}
\hline 
\multicolumn{1}{|c|}{\multirow{2}{*}{Methods}} & \multicolumn{2}{c|}{Test100}          & \multicolumn{2}{c|}{Rain100L}         & \multicolumn{2}{c|}{Rain100H}   & \multicolumn{1}{c|}{\multirow{2}{*}{Params (M)}}     \\ \cline{2-7} 
\multicolumn{1}{|c|}{}                        & \multicolumn{1}{c|}{PSNR}     & SSIM  & \multicolumn{1}{c|}{PSNR}     & SSIM  & \multicolumn{1}{c|}{PSNR}     & SSIM  &  \\ \hline \hline
\rowcolor{gray!25} Ideal                                         & \multicolumn{1}{c|}{$\infty$} & 1     & \multicolumn{1}{c|}{$\infty$} & 1     & \multicolumn{1}{c|}{$\infty$} & 1    & -- \\ \hline
HiNet~\cite{chen2021hinet}                   & \multicolumn{1}{c|}{30.29}    & 0.906 & \multicolumn{1}{c|}{37.28}    & {0.97}  & \multicolumn{1}{c|}{{30.65}}    & {0.894} & 88.7 \\ 
\rowcolor{gray!25} DerainNet~\cite{fu2017clearing}                & \multicolumn{1}{c|}{22.77}    & 0.810 & \multicolumn{1}{c|}{27.03}    & 0.884 &
\multicolumn{1}{c|}{14.92}    & 0.592 & -- \\ 
JORDER \cite{Yang_2017_CVPR}                & \multicolumn{1}{c|}{21.09}    & 0.753 & \multicolumn{1}{c|}{36.61}    & 0.974 &
\multicolumn{1}{c|}{26.54}    & 0.835 & -- \\ 
\rowcolor{gray!25} JORDER-E~\cite{yang2019joint}                  & \multicolumn{1}{c|}{27.08}    & 0.872 & \multicolumn{1}{c|}{37.10}    & 0.979 & \multicolumn{1}{c|}{24.54}    & 0.802 & 4.17  \\ 
SEMI~\cite{wei2019semi}                        & \multicolumn{1}{c|}{22.35}    & 0.788 & \multicolumn{1}{c|}{25.03}    & 0.842 & \multicolumn{1}{c|}{16.56}    & 0.486 & -- \\ 
\rowcolor{gray!25} DIDMDN~\cite{zhang2018density}                 & \multicolumn{1}{c|}{22.56}    & 0.818 & \multicolumn{1}{c|}{25.23}    & 0.741 & \multicolumn{1}{c|}{17.35}    & 0.524 & 0.372 \\ 
UMRL~\cite{yasarla2019uncertainty}             & \multicolumn{1}{c|}{24.41}    & 0.829 & \multicolumn{1}{c|}{29.18}    & 0.923 & \multicolumn{1}{c|}{26.01}    & 0.832 & 0.984 \\ 
\rowcolor{gray!25} RESCAN~\cite{li2016rain}                      & \multicolumn{1}{c|}{25.00}    & 0.835 & \multicolumn{1}{c|}{29.80}    & 0.881 & \multicolumn{1}{c|}{26.36}    & 0.786 & 0.150\\ 
PreNet~\cite{ren2019progressive}                     & \multicolumn{1}{c|}{24.81}    & {0.851} & \multicolumn{1}{c|}{32.44}    & 0.950 & \multicolumn{1}{c|}{26.77}    & {0.858} & 0.169 \\ \hline
%MSPFN~\cite{jiang2020multi}                  & \multicolumn{1}{c|}{\ul 27.50}    & {\ul 0.876} & \multicolumn{1}{c|}{32.40}    & 0.933 & \multicolumn{1}{c|}{28.66}    & {\ul 0.860} & -- \\ \hline
\rowcolor{gray!25} \textbf{DRT (Ours)}  & \multicolumn{1}{c|}{\textbf{27.02}}    & \textbf{0.847}  & \multicolumn{1}{c|}{\textbf{37.61}}    & \textbf{0.948}  & \multicolumn{1}{c|}{\textbf{29.47}}    &    \textbf{0.846}  & \textbf{1.18}  \\ \hline
\end{tabular}
%}

\label{Derain_Result_Table}
\end{center}
\end{table*}

We test DRT against other state-of-the-arts by comparing their derain results (PSNR and SSIM) on three different data sets along with the number of parameters and computational cost required. Then, an ablation study is carried out to investigate architecture variants. Lastly, we show that DRT can perform other image restoration tasks such as image desnowing.

\subsection{Datasets}
Experiments are carried out on the following data sets to evaluate the efficacy of the proposed DRT.

\noindent
\textbf{Rain800} \cite{zhang2019image}. This data set consists of 700 training images and 100 test images (Test 100). These images are chosen from the UCID data set \cite{schaefer2003ucid} and BSD500 \cite{arbelaez2010contour} training set with rain streaks synthesized on top of them. Different rain streaks are contained within Rain800. 

\noindent
\textbf{Rain100L} \cite{yang2017deep}. There are 1800 training images and 100 testing images in this data set. Background images are chosen from BSD200 data set \cite{martin2001database}. Each image consists of light rain streaks in one direction.

\noindent
\textbf{Rain100H} \cite{yang2017deep}. Similar to Rain100L, this data set consists of 1800 images and 100 testing images. Background are also chosen from BSD200 \cite{martin2001database}. However, each image can consist of heavy rain streaks in multiple directions, boosting deraining performances.

\noindent
\textbf{Snow 100K} \cite{liu2018desnownet}. Background images are downloaded from the Flickr API, and there are three types of snow densities (i.e., small, medium, and large) randomly selected and synthesized on the background images. The training set consists of 50K images. Three testing sets (i.e., Snow100K-S, Snow100K-M, Snow100K-L) correspond to three snow densities. Snow100K-S contains 16611 images, Snow100K-M consists of 16588 images, and Snow100K-L has 16801 images.

\subsection{Setup and Training}
\noindent
\textbf{Setup}. The best-performing model has the network structure shown in Figure 2. The patch embedding and image reconstruction stages consist of just one convolution layer without any activation. There are six RTBs (N = 6) in the deep feature extraction stage, and each of them consists of three recursive calls (L=3) on two STBs. Only a single convolution is employed at the end of each RTBs without any activation functions. All convolution operation maintains the input size, so there is no down-scaling or up-sampling of the image in the derain process that may cause the loss of pixel-level information. For each STB, the local window dimension is fixed to 7$\times$7, the patch size is set to one, and the number of heads is two. The depth of our hierarchical feature representations is chosen to be $D = 96$.
\begin{figure*}
\begin{center}
\begin{tabular}{@{}c@{}c@{}c@{}c}
\includegraphics[width=0.25\textwidth]{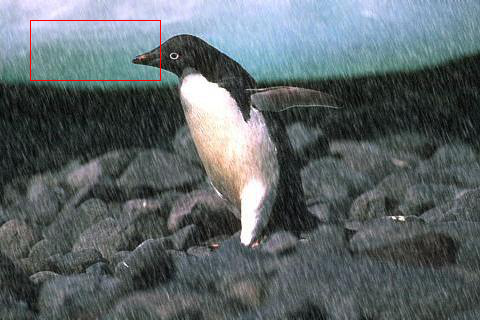} & \includegraphics[width=0.25\textwidth]{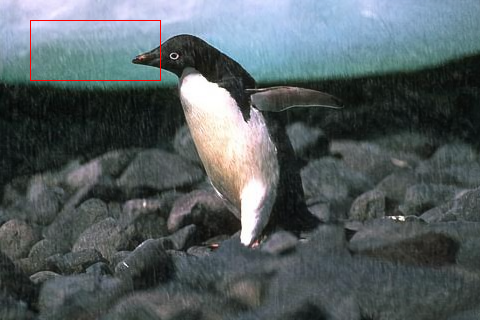} & \includegraphics[width=0.25\textwidth]{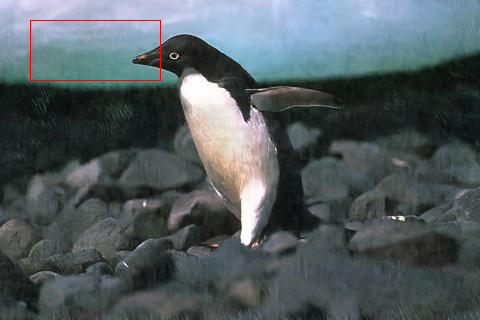} &
\includegraphics[width=0.25\textwidth]{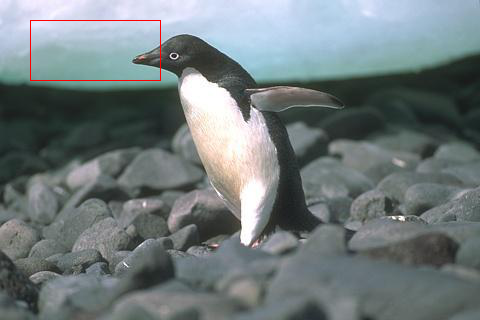} \\
\includegraphics[width=0.25\textwidth]{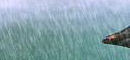} & \includegraphics[width=0.25\textwidth]{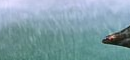} & \includegraphics[width=0.25\textwidth]{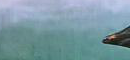} &
\includegraphics[width=0.25\textwidth]{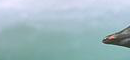} \\
Test100 Input & Hi-Net & DRT (Ours) & GT \\
% & Real Rain Input & DRT 
\end{tabular}
\caption{The image is drawn from Test100 data set. Very small rain streaks cannot fully removed by Hi-Net, however, our method is able to detect and remove these rain streaks.}
\label{fig:Test100}
\end{center}
\end{figure*}
\begin{figure*}
\begin{center}
\begin{tabular}{@{}c@{}c@{}c}
  \includegraphics[width=0.33\textwidth]{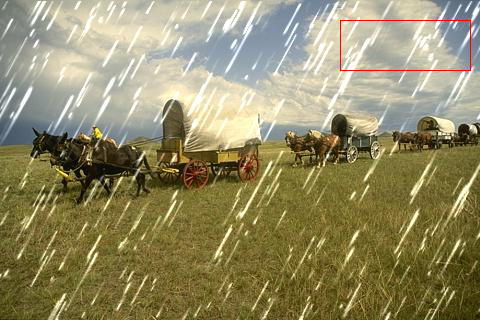} & \includegraphics[width=0.33\textwidth]{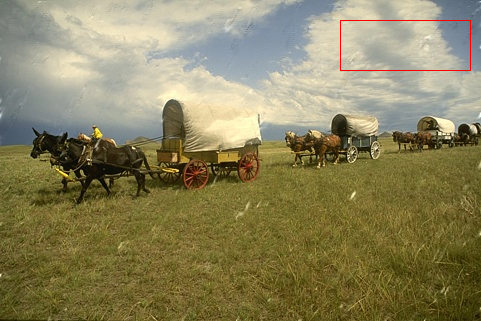}  & \includegraphics[width=0.33\textwidth]{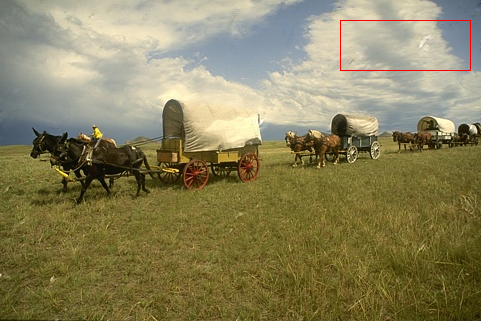} \\
  
  \includegraphics[width=0.33\textwidth]{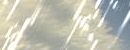} & \includegraphics[width=0.33\textwidth]{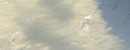}  & \includegraphics[width=0.33\textwidth]{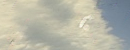} \\

  Input & JORDER & PreNet\\
  
  \includegraphics[width=0.33\textwidth]{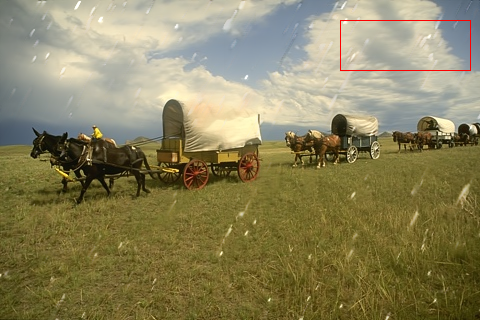} & \includegraphics[width=0.33\textwidth]{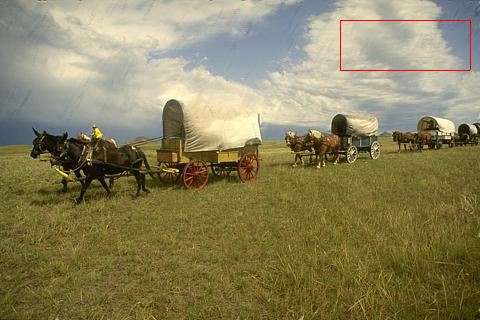} & \includegraphics[width=0.33\textwidth]{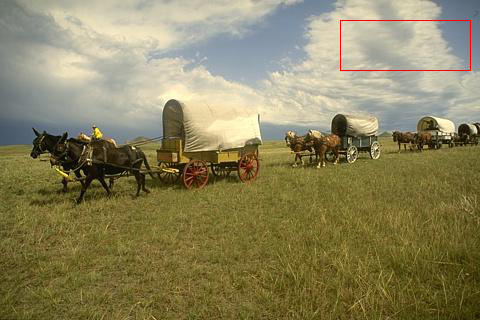}\\

  \includegraphics[width=0.33\textwidth]{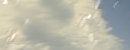} & \includegraphics[width=0.33\textwidth]{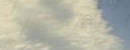} & \includegraphics[width=0.33\textwidth]{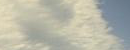}\\
  
  Hi-Net & DRT (Ours) & GT \\
\end{tabular}
\caption{Visual comparisons of derain results. This image is chosen from the Rain100L data set; the boxed region is enlarged and shown under each corresponding image. DRT can distinguish rain streaks from clouds in the first image, whereas the other methods cannot.}
\label{fig:Rain100L}
\end{center}
\end{figure*}

\noindent
\textbf{Training.} The initial training process is carried out mainly on the Rain-800 data set \cite{zhang2019image}, which consists of 700 synthetic-rain training samples with different types of rain streaks and 100 testing data (test100) for network validations. Instead of training it from scratch, we fine-tune this network on the other two data sets. The training process is slightly different for ablation studies, discussed in the following sections. A single RTX 2070 super graphics card with 8GB memory is used for the training. We use random cropping with a crop size, 56$\times$56, and random horizontal flipping on the training set to perform data augmentation. We choose the Adam optimizer for all training processes. The batch size is eight, and the initial learning rate is fixed to $1e^{-4}$. During the fine-tuning stage, the learning rate is adjusted to a value between $1e^{-5}$ and $1e^{-6}$. The network is trained until the error does not drop for 100 epochs, or the error only decreases by $1e^{-3}$ in the last 50 epochs. The best model we obtained is trained for 4600 epochs, and further training on this model leads to overfitting. Since many other transformer-based networks consume a large amount of data during training, we also tried a more extensive data set, Rain 13k \cite{jiang2020multi}, a mixture of different derain data set that contains a total of 13712 rainy-clean image pairs. However, we did not see any significant change in the performance of the networks. A small data set is sufficient for this model.

\subsection{Quantitative results}
The PSNRs and SSIMs of the best performing DRT model evaluated on three data sets and the results of comparisons with the other state of the arts are shown in Table~\ref{Derain_Result_Table}. Our results are in bold font for better visualization.

\begin{figure}[t]
    \centering
    \includegraphics[width=8cm]{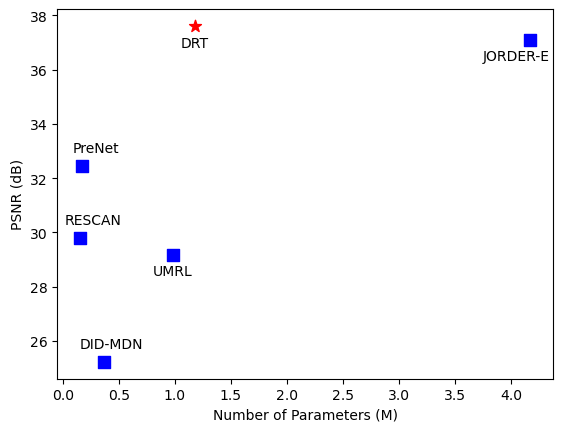}
    \caption{Number of Parameters vs. PSNR of different methods evaluated on the Rain100L data set. Methods locate towards the upper-left corner of the figure has better trade off between the number of parameters used and the derain performance. Models with large number of parameters are ignored in this graph.}
    %\vspace{-4mm}
    \label{fig:para vs psnr}
\end{figure}

For Rain100L, our performance exceeds all methods listed in Table~\ref{Derain_Result_Table} by at least 0.33db under the PSNR metric. For the same metric, our method's performances ranked second on the Test100 and Rain100H data sets. Similarly, DRT's results ranked second on the Rain100L under the SSIM metric. On the other hand, our method is very lightweight with respect to the number of parameters used, as seen in Table \ref{Derain_Result_Table}. Specifically, DRT only uses $\sim$ 1.3 \% of the number of parameters compared to HiNet. Note for early methods that did not release the source code; we omit their parameters in the table. In general, there is a trade-off between the number of parameters and the derain performance. Refer to Figure \ref{fig:para vs psnr} for a visualization of this trade-off. Models with a large number of parameters are ignored in this graph for better visualization. Methods closer to the upper-left corner tend to balance the number of parameters and the derain result and vice versa. DRT balances this trade-off well by being the closest to the upper-left corner, whereas methods such as Jorder-E and HiNet obtain good performance with the cost of large network sizes. Other methods may use fewer parameters but cannot achieve derain results as good as DRT. We further provide a fixed input to HiNet, PreNet, and DRT to calculate their corresponding multiplier-accumulator operations (MACs) (this metric is also used by HiNet) and the amount of memory needed to store the network and perform a forward and backward passing. The results are shown in Table \ref{Mac and Memory}. DRT consumes the least amount of MACs while enjoying roughly 4.5 times less memory usage than HiNet. 

\begin{table}[t]
\centering
\caption{MACs and memory usages of different methods. DRT (displayed in bold) consumes the least amount of MACs while still enjoying a very small memory usage.}
\resizebox{\linewidth}{!}{
\begin{tabular}{|c|c|c|c|c|}
\hline
Methods                      & \begin{tabular}[c]{@{}c@{}}Input Dim\\ (C$\times$H$\times$W)\end{tabular} & \begin{tabular}[c]{@{}c@{}}Input Size\\ (MB)\end{tabular} & \begin{tabular}[c]{@{}c@{}}MAC \\ (G)\end{tabular} & \begin{tabular}[c]{@{}c@{}}Size\\ (MB)\end{tabular} \\ \hline \hline
 \rowcolor{gray!25} HiNet                        & 3$\times$336$\times$336                                                 & 1.29                                                      & 293.79                                             & 2659.96                                             \\ 
PreNet & \multicolumn{1}{l|}{3$\times$336$\times$336}                              & 1.29                                                      & \multicolumn{1}{l|}{114.13}                        & \multicolumn{1}{l|}{417.96}                         \\ \hline
\rowcolor{gray!25} \textbf{DRT (Ours)}                  & \textbf{3$\times$336$\times$336}                                          & \textbf{1.29}                                             & \textbf{56.51}                                     & \textbf{587.19}                                     \\ \hline
\end{tabular}
}
\label{Mac and Memory}
\end{table}

%By carefully designing the recursive and residual structure, our network also does not rely on enormous training data sets while maintaining a good derain performance. 
%By carefully designing the recursive and residual structure, our network only uses 1.18M parameters and does not rely on enormous training data sets while maintaining a good derain performance. 

\subsection{Qualitative results}
\begin{figure}
\begin{center}
\begin{tabular}{@{}c@{}c@{}c}
\includegraphics[width=0.15\textwidth]{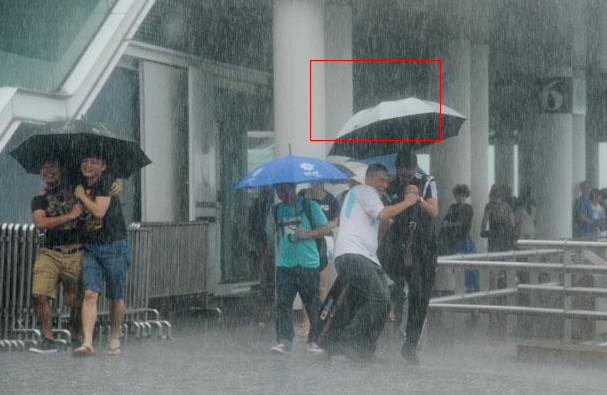} & \includegraphics[width=0.15\textwidth]{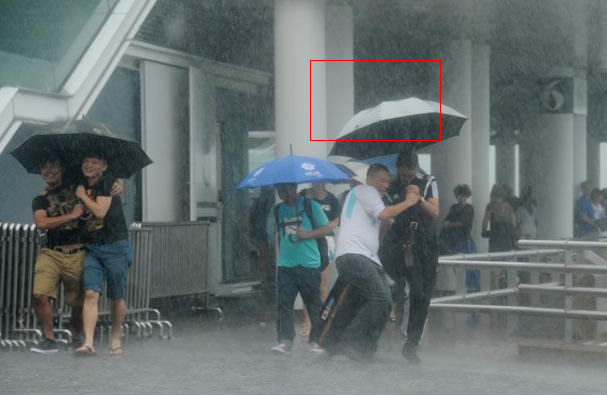} & \includegraphics[width=0.15\textwidth]{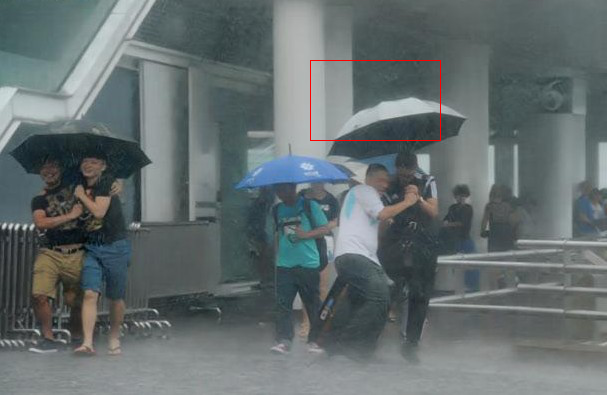} \\
\includegraphics[width=0.15\textwidth]{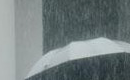} & \includegraphics[width=0.15\textwidth]{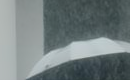} & \includegraphics[width=0.15\textwidth]{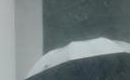} \\
Test100 Input & Hi-Net & DRT (Ours) \\
\end{tabular}
\caption{A realistic rainy image is presented here and DRT successfully removes all the rain streaks whereas Hi-Net ignores many of them.}
\label{fig:real_derain}
\end{center}
\end{figure}

We select visual outcomes of different methods evaluated on Test100, Rain100L and Rain100H and present them in Figure~\ref{fig:Test100}, Figure~\ref{fig:Rain100L} and Figure~\ref{fig:Rain100H} respectively. These images are chosen to show the deraining performance of different rain types. In Figure~\ref{fig:Test100}, the rain density is high, but each rain streak is very small. DRT is able to remove most of these rain streaks, whereas Hi-Net ignores many of them. Note that there is a tone shift in the background of the input and the background of the ground truth (GT) images. Since the training data do not consider background color changes, our restored image cannot recover the original tone. In Figure~\ref{fig:Rain100L}, other methods seem to interpret some rain streaks as parts of the land or the cloud and hence cannot fully remove these rains, whereas DRT can remove almost all of them. In Figure~\ref{fig:Rain100H}, we can see that artifacts are present in the smoke produced from the train, and artifacts exist in the smoke for images produced by Jorder and Hi-Net. Our model and PreNet give a better reconstruction of these complex soft-edge forms. Lastly, we also present a realistic derain outcome as seen in Figure~\ref{fig:real_derain}. Compared to Hi-Net, DRT can remove most of these rains, as seen in the zoomed-in pictures. Note for the realistic image that the haze caused by rains remains expected since the training data does not consider a mixture of haze and rain.

% \vspace{-2mm}

\subsection{Hyper-parameter Tuning}
We found out that the most critical design choice is reducing the patch size used by STBs. The original Swin transformer block sets the patch size to 4. We experience a significant increase in PSNR when reducing the patch size to one. This shows the importance of considering local pixel information in image deraining. Furthermore, we also found it beneficial to keep the input dimension (i.e., do not perform any up-sampling or down-sampling to the input) throughout the entire network. Since single image deraining is essentially a dense prediction task, these two design choices ensure there is no loss of pixel-level information for the network to infer. Other hyper-parameters of the model are also tuned extensively, and we summarise them here. As mentioned before, we did not find any impact on the use of the shift-window approach from the original Swin transformer; this is likely due to the fact that we have used convolutions at the end of each RTB and local positional encoding for each STB to consider both local and global positional information whereas the shift-window approach is proposed to solve the loss of global information in the original transformer. Similarly, we tried to increase the depth of the transformer's working dimension from 96 to 180, but there is no improvement in performance. We tried to drop the residual connection between the input and the output, but no improvement was observed. This is likely because residual connections limit the inference stage to detect rain streaks, which occupy a small portion of information in the input image. 

\begin{table}
\caption{Ablation Studies of DRT Variants on Rain100L. The model still performs well when reducing the number of STBs in each RTB to one while using fewer parameters (eighth row). Without any recursion, the performance drops dramatically. The best configuration is displayed in bold.}
\centering
%\resizebox{\linewidth}{!}{
\begin{tabular}{|c|c|c|c|c|} 
\hline
\begin{tabular}[c]{@{}c@{}}N\end{tabular} & \begin{tabular}[c]{@{}c@{}}L\end{tabular} & \begin{tabular}[c]{@{}c@{}}STBs \end{tabular} & \begin{tabular}[c]{@{}c@{}c@{}}Params (M) \end{tabular} & PSNR  \\ 
\hline \hline
\rowcolor{gray!25} 3                                                                      & 3                                                                           & 2                                                                     & 0.591                                                                       & 10.93          \\ 

9                                                                      & 3                                                                           & 2                                                                     & 1.77                                                                        & 10.25          \\ 

\rowcolor{gray!25} 6                                                                      & 1                                                                           & 2                                                                     & 1.18                                                                        & 6.072          \\ 

6                                                                      & 2                                                                           & 2                                                                     & 1.18                                                                        & 19.61          \\ 

\rowcolor{gray!25} 6                                                                      & 4                                                                           & 2                                                                     & 1.18                                                                        & 24.08          \\ 

8                                                                      & 2                                                                           & 2                                                                     & 1.57                                                                        & 14.88          \\ 

\rowcolor{gray!25} 6                                                                      & 3                                                                           & 1                                                                     & 0.841                                                                       & 35.55          \\ 

6                                                                      & 3                                                                           & 3                                                                     & 1.52                                                                        & 36.96          \\
\hline
\rowcolor{gray!25} \textbf{6}                                                                      & \textbf{2}                                                                           & \textbf{3}                                                                     & \textbf{1.18}                                                                        & \textbf{37.61}          \\
\hline
\end{tabular}
%}
\label{Ablation1}
%\vspace{-4mm}
\end{table}
% \vspace{-2mm}

\subsection{Ablation Studies}
We then perform an ablation study on the essential design features of the DRT, namely, the number of RTBs ($N$), the number of STBs, and the number of recursive calls on STBs ($L$). During this study, each model is trained on Rain100L from scratch with a similar training procedure. All other hyper-parameters not included in the ablation study are fixed on the optimal network structure. The evaluation of each model is performed on the Rain100L data set. The number of parameters and the PSNR for each evaluation is listed in Table \ref{Ablation1} with the best model displayed in the last row. As seen in the table, the first two rows study the impact of the number of RTBs concerning the best DRT model; however, corresponding low PSNR results show that both increasing and decreasing the number of RTBs dramatically limits the performance of DRT. Row three to five studies the influence of the number of recursive calls of STBs on the performance of DRT. Once again, neither increasing nor decreasing the number of calls can boost the derain performance of DRT. Note that having only one recursive call refers to a variant of DRT without any recursion. However, its performance is the worst among all other variants, which shows the importance of recursions in the design. We then tune both the number of RTBs and the number of recursive calls simultaneously, but the result is not good. Hence, we conclude that having six RTBs and three recursive calls is essential for a promising derain performance. Contrasting to this, the second and third last row of Table \ref{Ablation1} indicate, that changes in the number of STBs do not influence the derain result much. It is worth noting that having just one STB lowers the number of parameters from 1.18M to below a million, 0.841M, while still keeping a good derain outcome.

\begin{table}[t]
\caption{PSNR results of different methods on Snow100K. Our results showing in bold consistently perform best or second best.}
\centering
\resizebox{\linewidth}{!}{
\small
\begin{tabular}{|l|c|c|c|}
\hline
Methods                                                           & Snow100K-S & Snow100K-M & Snow100K-L \\ \hline \hline
\rowcolor{gray!25}DehazeNet \cite{cai2016dehazenet}& 24.96               & 24.16               & 22.62               \\ 
DeepLab \cite{chen2017deeplab}    & 25.95               & 24.36               & 21.29               \\ 
\rowcolor{gray!25}DesnowNet \cite{liu2018desnownet} & {32.33} & {30.86} & {27.17} \\ \hline
%\begin{tabular}[c]{@{}l@{}}AllinOne \cite{li2020all}\end{tabular}        & ---                 & ---                 & \textbf{28.33}      \\ \hline
\textbf{DRT (Ours)}                                                                        & \textbf{32.15}         & \textbf{31.20}      & \textbf{28.04}         \\ \hline
\end{tabular}
}
\label{desnow-table}
\end{table}

\begin{figure}[t]
\begin{center}
\begin{tabular}[h]{@{}c@{}c}
\includegraphics[width=0.23\textwidth]{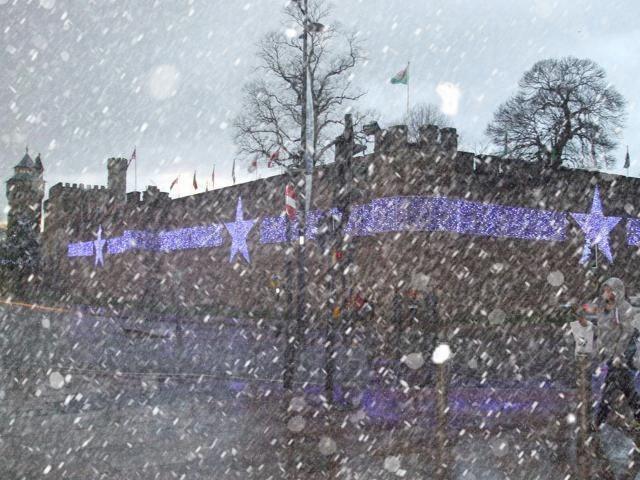} & \includegraphics[width=0.23\textwidth]{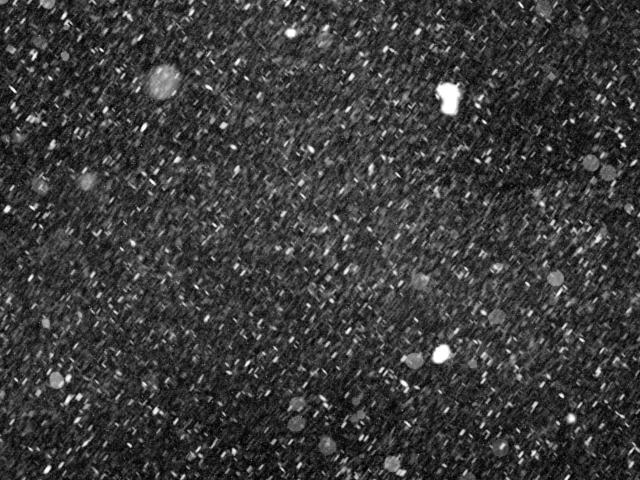}\\
Input & Snow Mask \\
\includegraphics[width=0.23\textwidth]{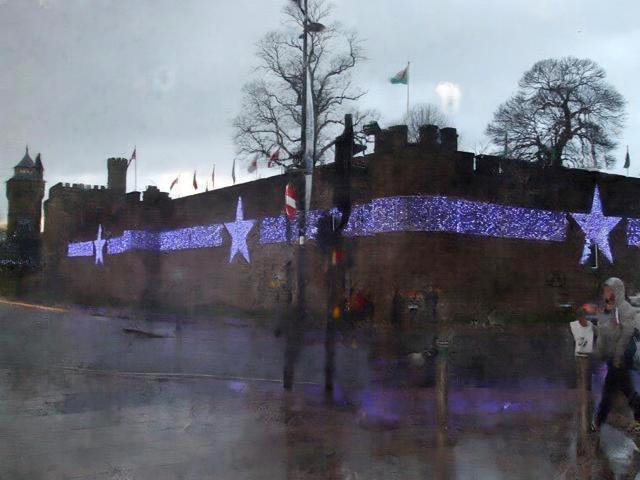} & \includegraphics[width=0.23\textwidth]{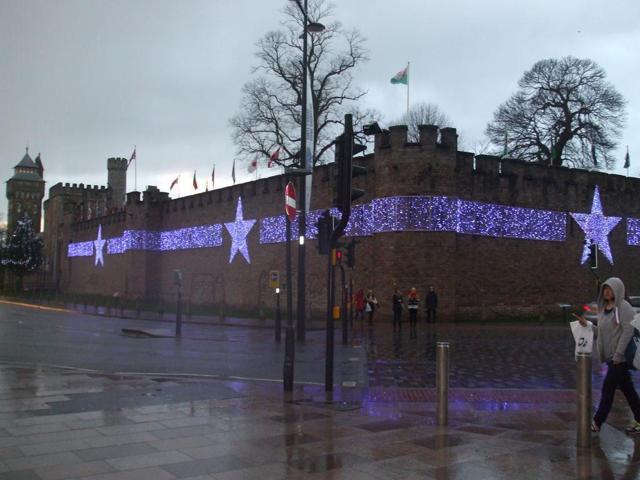}\\
Ours & GT
\end{tabular}
\caption{A sample visual result drawn from the Snow100K-L. As seen in the input and mask images that snows are characterised by different underlying functions with respect to rain streaks. DRT successfully removes mixture of dense snows from the input.}
\label{fig:desnow}
\end{center}
%\vspace{-8mm}
\end{figure}

\subsection{Results on Single Image Desnowing}
We further test DRT's performance on the task of single image desnowing to empirically show its capability in solving different image restoration tasks. The best model is trained on the Snow100K data set \cite{liu2018desnownet} with the same training procedure as before. The model is then tested on the corresponding three data sets provided by Snow100K; each consists of one snow type. The PSNR results of different methods are shown in Table \ref{desnow-table}. Our model's performance ranks either the first or the second among these methods. We also display a sample of desnow visual results with the corresponding input, output, and mask images in Figure~\ref{fig:desnow}. Although the snow in the input image is dense, DRT can remove most of them and restore a clean background that is very close to the ground truth. Rain streaks and snowflakes can be very different both physically and visually, which means they require different functions to be removed; this study shows that upon training, DRT is not restricted to single image deraining but can also be well applied to other fields.

\section{Conclusion}
Transformer-based networks are often computationally heavy. We utilize the local window-based self-attention mechanism, the residual, and the recursive connection to achieve rain streak removal with a lightweight architecture for image deraining. It performs the best on the Rain100L data set and maintains top performances for other data sets. Due to its simple structure, it uses a small number of parameters while relying on less computing resources. Since there is no deliberate design for removing rain streaks in the model, we show that it can be applied to desnowing and potentially to other image reconstruction tasks.

%%%%%%%%% REFERENCES
{\small
\bibliographystyle{ieee_fullname}
\bibliography{refs}
}

\end{document}